# Neural Network Layer Matrix Decomposition reveals Latent Manifold Encoding and Memory Capacity

Ng Shyh-Chang[1,2,3,5], A-Li Luo[4,5], Bo Qiu[6]

[1]Institute of Zoology, Chinese Academy of Sciences, Beijing, China.

[2]Institute for Stem Cell and Regeneration, Chinese Academy of Sciences, Beijing, China.

[3]Beijing Institute for Stem Cell and Regenerative Medicine, Beijing, China.

[4]Key Laboratory of Optical Astronomy, National Astronomical Observatories, Chinese Academy of Sciences, Beijing, China

[5]University of Chinese Academy of Sciences, Beijing, China

[6]School of Electronic and Information Engineering, Hebei University of Technology, Tianjin, China

huangsq@ioz.ac.cn , lal@nao.cas.cn , qiubo@hebut.edu.cn

**Abstract:**

We prove the converse of the universal approximation theorem, i.e. a neural network (NN) encoding theorem which shows that for every stably converged NN of continuous activation functions, its weight matrix actually encodes a continuous function that approximates its training dataset to within a finite margin of error over a bounded domain. We further show that using the Eckart-Young theorem for truncated singular value decomposition of the weight matrix for every NN layer, we can illuminate the nature of the latent space manifold of the training dataset encoded and represented by every NN layer, and the geometric nature of the mathematical operations performed by each NN layer. Our results have implications for understanding how NNs break the curse of dimensionality by harnessing memory capacity for expressivity, and that the two are complementary. This Layer Matrix Decomposition (LMD) further suggests a close relationship between eigen-decomposition of NN layers and the latest advances in conceptualizations of Hopfield networks and Transformer NN models.

**Introduction**

Each layer of a Neural Network (NN) can be represented as a matrix of weights acting on a vector of input data or activation functions. Specifically, the input to each layer is a vector of activation values from the previous layer, which is multiplied by a weight matrix to produce a new vector of weighted inputs. At each neuron, the weighted inputs are then passed through an activation function to produce the output of the layer.

Consider a feedforward NN [1] of depth L, with $l^{th}$ layer width $n_l \leq L$. Let $W^l_{ij}$ be the weight matrix connecting layer *l-1* to layer *l* (absorb all bias terms $b_l$ into additional rows of the weight matrix $W^l_{ij}$). Let X be the input space, Y be the target space. Let D = {$(x_i, y_i)$} be a representative training dataset with $x_i \in X$, $y_i \in Y$. For some input vector $a \in R^n$, the propagation of this input through the NN is given for activation functions $\varphi : R \rightarrow R$ by

$$y^l_i(a) = \sum^{Nl-1} W^l_{ij} \varphi(y^{l-1}_j(a)) \quad (1)$$

As is common to the NN literature [1], we assume:

1) X is a compact metric space with distance function dX.

2) Activation functions $\varphi(z)$ are continuous and bounded on their domain (e.g. ReLu, tanh).

3) The loss function L(W) is continuous and differentiable in W.

4) The NN trained on D converges stably: $\|W^* - W\| \leq M^* \|\nabla L(W^*)\|$ for some $M^* > 0$.

**Theorem 1 (Existence of encoded memories):** *After a NN converges, the optimized weights $W^{l}_{ij}{}^*$ have encoded a continuous function that approximates D within a finite margin of error $\varepsilon$.*

Proof: By assumptions 1 and 2, each layer $l$ computes a continuous function of the previous layer. Composing the L layers, $NN(x;W)$ is a continuous function $f_W(x)$ of x. By the Heine-Cantor theorem, as X is compact (assumption 1), $f_W$ is uniformly continuous on X.

Since $W^*$ minimizes $L(W)$ by assumptions 3 and 4, $f_{W^*}$ approximates D:
$$f_{W^*}(x_i) \approx y_i, \forall (x_i, y_i) \in D.$$
Thus, $\exists \delta > 0$ such that $d_X(x_1, x_2) < \delta \Rightarrow |f_{W^*}(x_1) - f_{W^*}(x_2)| < \varepsilon$.

By assumptions 3 and 4, small dW results in small dL. Specifically,
$$\|W - W^*\| < M^* \|\nabla L(W^*)\| \Rightarrow |L(W) - L(W^*)| < \varepsilon.$$

Let $W'$ satisfy $\|W' - W^*\| < \min(\delta, \varepsilon/M^*)$.
Then $|f_{W'}(x) - f_{W^*}(x)| < \varepsilon \; \forall x \in X$.

Therefore, the converged weights $W^*$ encoded or memorized a continuous function $f = f_{W^*}$ that approximates D within error $\varepsilon$.

To achieve approximation of a high-dimensional function *f* with generality (within error margin $\varepsilon$) and break the [curse of dimensionality](#) [1], a minimum memory capacity within $W^*$ is a necessity and prerequisite. It is a common fallacy to assume that simply memorizing data will inevitably lead to overfitting, especially in the presence of noise. However, we argue against this fallacy by pointing out the existence of neural network (NN) regularization, both implicit and explicit, in order to suppress the influence of noise and achieve stable convergence of the loss function $L(W)$. Moreover, our definition of function encoding applies not only to NNs that achieved stable convergence but also only to a truly representative dataset D.

To establish conditions for approximation of *f* with generality, we need both necessary and sufficient conditions to break the curse of dimensionality. Firstly, a minimum memory capacity within $W^*$ for D is necessary to capture more *n* samples of representative training data, which [allows us to localize to fewer function possibilities, and break the curse of dimensionality [1]](#). Secondly, if we can effectively utilize the memory capacity for [hierarchical/compositional representations](#) [2] in deep networks with large L to iteratively [reduce input space dimensions with each layer [3]](#), also implied by the Kolmogorov-Arnold Representation theorem (as a solution to Hilbert's $13^{th}$ Problem), we can further break the curse of dimensionality.

In summary, a minimum memory capacity in the $W^*$ and thus NN is essential for both (i) capturing more representative training data and localizing to fewer function possibilities, and (ii) utilizing deep networks' memory capacity for hierarchical/compositional representations and achieving dimensionality reduction. With the help of memory capacity, such that these 2 conditions are fulfilled as well, we can go from necessity to sufficiency conditions for NN-based approximation of *f* with generality.

***Theorem 2 (Geometry of encoded memories):*** *Every NN layer encodes a latent representation of the D manifold after stable convergence.*

Proof: From Eqn 1 and Theorem 1, the NN $l^{th}$ layer's activation function output vector $\varphi(y^{l-1})$ multiplied by the weighted adjacency matrix $W^l_{ij}$ encodes part of the function *f* mapping the *i*-dimensional input manifold to the *o*-dimensional output manifold $f: R^i \rightarrow R^o$. We will now show that a singular value decomposition (SVD) of this matrix $W^l_{ij}$ can illuminate the location of the memories encoded by the NN in each layer, and the geometric nature of the mathematical operations performed by each NN layer. According to the SVD Theorem, every m × n matrix can be factorized into

$$M = U \times S \times V^T \quad (2)$$

where U is an m × m orthogonal matrix, S is an m × n diagonal matrix with non-negative real numbers on the diagonal, and V is an n × n orthogonal matrix. Diagonal elements of S, known as the singular values (or rooted eigenvalues of $M \times M^T$), are arranged in descending order.

Consider the [Eckart–Young-Minsky theorem for truncated SVD](), if we just use the *n'* largest singular values and first *n'* columns of S, SVD will give us a lower rank *n'* matrix M' approximation of M, such that S' is now a *n' × n'* diagonal matrix, containing the *n'* largest eigenvalues of M. As a result of these 2 rounds of SVD, we can construct a linear approximation of the weight matrix $W^l_{ij}$ by factorizing it into:

$$W^l_{ij} = U \times O_{dist} \times S' \times I_{dist} \times V^T \quad (3)$$

For an i × o $W^l_{ij}$ matrix that maps a vector of dimension i to a vector of dimension o, i.e. approximating an $f': R^i \rightarrow R^o$, where U and V are orthogonal matrices of dimensions o × o and i × i respectively, S' is an *n' × n'* diagonal matrix with the largest *n'* singular values of $W^l_{ij}$ in descending order, $O_{dist}$ and $I_{dist}$ are o × *n'* and *n'* × i rectangular matrices. Next we will show that S' represents the *n'*-dimensional latent space of *f*, and $O_{dist}$ and $I_{dist}$ are constructed to map the metric transformations between the output/input spaces and the *n'*-dimensional latent space.

We can interpret U and $V^T$ as isometric transformation matrices, e.g. distance-preserving rotations, reflections, and translations in the case of Euclidean spaces. In each NN layer, these correspond to the weighted linear transformation we perform on the input vector, before and after each neuron's sigmoidal or (non)linear activation function acts on it. In contrast, we can interpret S' in the truncated SVD of S in $M=U \times S \times V^T$ as a linear approximation of the latent space within the mapping $f: R^i \rightarrow R^{n'} \rightarrow R^o$ represented by the $W^l_{ij}$ matrix. Such a linear approximation to compress the $R^i \times R^o$ manifold representing *f* into a lower n'-dimensional subspace or latent space manifold, is possible according to the manifold learning hypothesis.

In this vein, the truncated SVD approximation of S allows us to directly obtain the *n'*-dimensional latent space, represented by the *n' × n'* diagonal matrix of eigenvalues S', where 0 < n' < i. We can then construct metric transformation maps $I_{dist}$ and $O_{dist}$, between U, $V^T$ and S' as such:

$$U \times O_{dist} \times S' \times I_{dist} \times V^T = U \times S \times V^T$$
$$O_{dist} \times S' \times I_{dist} = S \quad (4)$$

This means I/$O_{dist}$ is just the identity matrix for the first *n'* rows and columns, and zero or degenerate everywhere else. This is the most trivial case of a metric transformation, where the unimportant dimensions are simply collapsed to zero, and the inner product is now only defined over the non-zero dimensions. This extreme case applies when the latent manifold has a topology equivalent to the input/output manifolds' topologies.

We can further generalize this idea, if we assume the latent manifold has a different topology from the input/output manifolds. We will use graphs to discretize the manifolds. In essence, the truncated SVD assumes an equivalent graph topology between the dataset XY's (input X × output Y) manifold and the latent manifold, i.e. every data point $\{(x_i, y_i)\}$ need only be represented by their coordinates, and thus a fully-connected product graph can be constructed between the input X graph and output Y graph, and between the $p$ data points in $R^i \times R^o$ space [4]. However, if there is a different topology between X, Y and the latent manifold, then every data point $\{(x_i, y_i)\}$ in the original $R^i \times R^o$ space is actually not equally connected in the latent space, and we can construct a locally-connected graph between the $p$ data points (weighted by unequal adjacency) as a discretization of the original XY manifold. Then we can define the $p \times p$ graph Laplacian matrix L as follows:

$$L = D - A \quad (5)$$

where D is the degree matrix, A is the adjacency matrix

In the case of the fully-connected graph, one can prove that solving the [eigendecomposition of L and taking the largest $n'$ eigenvectors of L is equivalent to solving for the truncated SVD](#) of XY to get S' [4]. Similarly, in the case of the locally-connected graph, one can prove that the L matrix is equivalent to a discretization of the [Laplace-Beltrami operator](#) on the $(i + o)$-dimensional manifold defined by XY, such that L's eigen-decomposition allows us to obtain a $n'$-dimensional latent space manifold to define S'. In this dimensional reduction based on locally-connected graph representations of XY, the I/O$_{dist}$ metric transformation matrices are no longer identity matrices for the first $n'$ rows and columns. Instead, they can be theoretically calculated by:

$$I_{dist} = c(L \times L_{in}^{-1})$$
$$O_{dist} = c'(L_{out} \times L^{-1}) \quad (6)$$

where L is the graph Laplacian matrix for the latent manifold, $L_{in}$ and $L_{out}$ are the Laplacian matrices for the input graph X and output graph Y respectively, $L^{-1}$ is the pseudo-inverse matrix of L (and where the c, c' matrices are merely used to reshape the graph Laplacian matrices for $n'$, $i$, $o$ dimensions via hypergraph spectral clustering, another eigen-decomposition), such that if we denote the I/O$_{dist}$ matrices as a mapping function $g$: metric$_A$ → metric$_B$, it satisfies

$$\text{metric}_B = \text{integral } g(\text{derivative of metric}_A) = \text{integral } g(L_A)$$
$$\text{integral}(L_B) = \text{integral } g(L_A)$$
$$L_B = g(L_A); \text{ so } g = L_B \times L_A^{-1} \quad (7)$$

Note: in truncated SVD, $g$ is an identity matrix (Eqn 4), implying $L_B$ and $L_A$ are equivalent, and their corresponding graphs are equivalent in topology.

Multiplying a vector by the Laplacian matrix of a graph has a geometric interpretation related to the diffusion process on the graph. The Laplacian matrix L can be seen as a discrete approximation of the Laplace-Beltrami operator on a manifold, and it models the diffusion of a signal on the graph over time. Specifically, starting with an initial signal represented by a vector $x$, multiplication of L and $x$ yields L$x$, which represents the spread of the signal on the graph after one time step. This process can be repeated iteratively to simulate the diffusion of the signal over multiple time steps.

Returning to the SVD factorization of each NN layer, as framed in graph theory, we have:

$$Y = Wx = USV^Tx = UO_{dist}S'I_{dist}V^Tx = Uc'L_{out}L^{-1}S'cLL_{in}^{-1}V^Tx \quad (8)$$

We call Eqn 8 a Layer Matrix Decomposition (LMD) of NNs. Let $Y=Wx$ be a stably converged large (or infinite) width single-layer NN, and $x$ be an input signal vector. Then we can interpret $V^T$ as the first isometric transformation on $x$, $I_{dist} = c*L*L_{in}^{-1}$ as a metric transformation from the input manifold to the latent manifold (via reversed diffusion then re-diffusion of the signal on their respective graphs) to produce $x'$, S' as the latent mapping function/manifold that maps $x'$ to an approximated $y'$ in latent space, $O_{dist} = c'*L_{out}*L^{-1}$ as another graph-based metric transformation from the latent manifold to the output manifold, and U as the second isometric transformation.

When the activation functions are ReLU's (or other sigmoidal functions that can be approximated with linear units like ReLU's), the $n' \times n'$-dimensional S' matrix constitutes the minimal series of ReLU's that can $n'$-piecewise approximate and solve for the $n'$ dimensions of the latent manifold. We can extend this logic to assume that $Y = Wx$ is a single layer within a stably converged DNN, and $x$ is taking the output from the previous $l^{th}$ layer as the input. In this case, each layer could potentially approximate the target function $f$ progressively, layer-by-layer. Note that this interpretation of LMD does not assume any layer-by-layer increase or decrease in approximation accuracy or latent space dimension, either scenario is possible.

Computationally, LMD obtains the $n'$-dimensional latent space S' by computing the diagonal matrix of the $R^i \times R^o$ data space with SVD and learning the eigen-decomposible graph Laplacian matrices of the input, latent, and output manifolds. Specifically, we have:
$$S' = I_{dist}^{-1} \times S \times O_{dist}^{-1}$$
Geometrically, this would be equivalent to reshaping the eigenvector space of S with a series of "curvature" metric transformations, according to the topological characteristics of the respective graph Laplacians.

Suppose $x$ is an $i$-dimensional input vector (from the training dataset), re-oriented ($V^T$) and reshaped to the minimal/latent dimension of $0 < n' < i$ with the right metric ($I_{dist}$) to form $x'$, $S'x'$ will give the correct re-scaling of $x'$ on each dimension (e.g. one ReLu per dimension) to produce the correct $y'$ of dimension $n'$, that will lead to the correct $y$ (from the training dataset) of dimension $o$ after another metric transformation ($O_{dist}$) and re-orientation (U). "Correct" is defined as being accurate within the margin of error ε in Theorem 1. In this way, all $p$ elements of the training dataset of $\{(x_i,y_i)\}$ have been encoded within the latent manifold/mapping function represented by S'. Thus, after stable convergence of the NN (which is guaranteed by Theorem 1), S' holds a Hopfield-like "memory" of the training dataset in the form of a latent mapping function (auto-associative if $y=x$ or hetero-associative if $y \neq x$), within a margin of error ε.

Alternatively, we can use an $i$-dimensional vector of function coefficients for an $o$-dimensional output vector of function coefficients, if we prefer not to use ReLUs to approximate the manifold and prefer to use nonlinear activation functions instead. These activation functions can be regarded as kernel functions, and we would be implicitly doing a complex combination of kernel PCA with metric transformations (according to Eqn 8), in a higher-dimensional space. This approach may make it easier to separate and classify the data.

In general, each of the matrices in this LMD factorization of a transition weight matrix could either be a trivial "identity" layer or an aggregate of many layers in an actual NN, and either $i > o$ or $i < o$ could hold. Thus at least some layers in a stably converged NN progressively "reshapes" the original data manifold into a composite series of latent manifolds, encoding/memorizing the training dataset of $\{(x_i, y_i)\}$ as a series of eigen-decomposible mapping functions in latent spaces, to enable an accurate, generalizable, and speedy processing of (potentially complex and high-dimensional) input-output signals.

**Previous Works**:

Our LMD factorization delves into the mathematical underpinnings of the memorization/encoding phenomenon in NNs and establishes connections to recent formulations of Universal Hopfield Networks (UHNs) [5], Associative Memory Networks, and Transformers [6-8]. In the UHN framework proposed by Millidge et al. [5], a single-shot associative memory is viewed as a function that takes an input vector q, ideally a corrupted version of a data point already in memory, and outputs a vector corresponding to the closest stored data point. Their UHN framework shows that every feedforward associative memory in the literature admits the following factorization, which defines an abstract and general Hopfield network:

$$z = Proj * Sep(Sim(M, q)) \quad (9)$$

where $z$ is the output vector of the memory system, *Proj* is a projection matrix, *Sep* is the separation function, *Sim* is the similarity function, $M$ is a matrix of stored memories or data points, and $q$ is the input query vector.

The intuition behind this UHN factorization (Eqn 9) is to rank how similar an input query is to all other stored memories and retrieve the closest one [5]. This is achieved by the Similarity function, which outputs a vector of similarity scores between each data point held in the memory and the input query. The Separation function then emphasizes the top score and de-emphasizes the rest to produce a clear output pattern. They showed that separation functions of higher polynomial degrees lead to capacity increases of the order of $C \sim N^{(n-1)}$, where N is the number of visible (input) neurons, and n is the order of the polynomial [8-12]. Exponential separation functions, such as the softmax, lead to exponential memory capacity. However, they experimentally showed that the real bound on NN performance is not just memory capacity but rather the ability of the Similarity function to distinguish between the input query and various possible stored patterns [5].

In our LMD factorization, we provide a more mathematically grounded formulation of the transition weight matrix of every NN, where the Similarity function is analogous to the $I_{dist}$ metric transformation matrix based on the graph Laplacians, instead of using a trial-and-error approach with different metrics, from dot product to Manhattan distance [5]. The Separation function is analogous to the latent manifold/mapping function S', especially when it is acting on a vector of kernel functions or nonlinear activation functions. Modern continuous Hopfield networks [6] and Transformer network models, such as BERT and GPT, use the exponential softmax function, while Sparse distributive memories (SDMs) use the linear thresholded function, and classical Hopfield networks use the trivial identity function [5]. The Projection matrix is analogous to the U*$O_{dist}$ matrix product which transforms the latent manifold to the output manifold. Moreover, our LMD factorization admits a natural symmetry intrinsic to the matrix eigen-decomposition of graphs and provides a computational algorithm based on the intrinsic topology/geometry of the dataset.

Importantly, our LMD factorization applies not just to Associative Memory models, but to all NNs. Through this layer-by-layer factorization, we can calculate the latent space manifold S' that encodes the NN's memories (instead of an abstract M matrix) and calculate how the memories are being transformed layer-by-layer. We can further illuminate the relationship between memory capacity and the dimensionality of S', and how kernel functions of $n^{th}$ degree expands this memory capacity by the power of n-1, as an alternative to the abstract energy landscape interpretation of Associative Memory capacity. Briefly, by the Stone-Weierstrass approximation theorem and Taylor's series theorem, every nonlinear continuous function on a closed interval can be approximated with a higher degree of precision with a higher order polynomial kernel function (with $e^x$ or softmax being an extreme case, as it cannot be uniformly approximated by a polynomial of finite degree to arbitrary precision). Thus the use of $n^{th}$ order polynomials (where n≥1) as activation functions raises all manifolds' (input, output, latent) dimensions by the power of n-1, therefore increasing the memory capacity of the S' latent manifold by the power of n-1. When we use $e^x$, we increase the memory capacity infinitely. But these actions also increase the *p* elements needed to approximate the S' manifold accurately, which would only make sense if we have a very large dataset. It also explains why Transformer models work so well, because their "attention" mechanism directly calculates the dot product distance from the "Key" vectors for the $I_{dist}$ matrix, and uses the high memory capacity $e^x$/softmax function for kernel PCA of S' in the context of a very large dataset. Our LMD factorization would further suggest that modifications to the architecture to directly calculate the U, $O_{dist}$, S', $I_{dist}$, $V^T$ matrices and estimate the kernel function degree needed, based on the dataset's size and topology instead of arbitrary assumptions, would further improve the performance of NNs. In future work, it would also be interesting to extend the Lyapunov energy vs time dynamics perspective on Hopfield memory network convergence to our LMD factorization of NN layers and examine how we can optimize NN performance based on the discrete calculus of graph matrices.

**Conclusion**
The hierarchical and modular nature of this LMD factorization merits further analysis as a model of how simple functions become complex non-linear functions after assembling into networks. Explicit incorporation of the factorization process into neural network architectures might also enable optimization based on the intrinsic dimensionality and geometry of data manifolds. For example, meta-learned topology-aware weight initializations, dynamic dimensionality reduction guided by singular values, and direct regularization of the metric transformations could improve NN efficiency and generalization. Exploration of these techniques remains an open problem with promising potential.

In summary, our LMD factorization elucidates the implicit dimensionality reduction and manifold learning occurring within neural network computation. This geometric view enables analysis of each NN model's memory capacity and suggests techniques to improve efficiency based on the topology of data. Connecting to dynamical systems theory and biological networks offers exciting directions for further understanding the universal learning dynamics in both artificial and natural networks of simple agents.

**References**

1. Martin Anthony and Peter L. Bartlett. Neural Network Learning: Theoretical Foundations. Cambridge University Press, NY, NY, USA, 1999.
2. Tomaso Poggio, Hrushikesh Mhaskar, Lorenzo Rosasco, Brando Miranda, Qianli Liao. Why and When Can Deep -- but Not Shallow -- Networks Avoid the Curse of Dimensionality: a Review. arXiv: 1611.00740v5, 2017.
3. T. Poggio, F. Anselmi, L. Rosasco. I-Theory on Depth vs Width: Hierarchical Function Composition. Center for Brains, Minds and Machines [CBMM], Cambridge, MA, 2015.
4. Andrew Draganov. Towards a Common Dimensionality Reduction Approach; Unifying PCA, tSNE, and UMAP through a Cohesive Framework. http://hdl.handle.net/1920/12149
5. Beren Millidge, Tommaso Salvatori, Yuhang Song, Thomas Lukasiewicz, Rafal Bogacz. Universal Hopfield Networks: A General Framework for Single-Shot Associative Memory Models. arxiv: 2202.04557v2, 2022.
6. Ramsauer, H., Schafl, B., Lehner, J., Seidl, P., Widrich, M., Adler, T., et al. Hopfield networks is all you need. arXiv: 2008.02217, 2020.
7. Krotov, D., & Hopfield, J. Large associative memory problem in neurobiology and machine learning. arXiv: 2008.06996, 2020.
8. Krotov, D., & Hopfield, J. Dense associative memory for pattern recognition. Advances in Neural Information Processing Systems, 29, 1172–1180, 2016.
9. Abbott, L. F., & Arian, Y. Storage capacity of generalized networks. Physical Review A, 36(10), 5091, 1987.
10. Baldi, P., & Venkatesh, S. S. Number of stable points for spin-glasses and neural networks of higher orders. Physical Review Letters, 58(9), 913, 1987.
11. Caputo, B., & Niemann, H. Storage capacity of kernel associative memories. In International conference on artificial neural networks (pp. 51–56), 2002.
12. Chen, H., Lee, Y., Sun, G., Lee, H., Maxwell, T., & Giles, C. L. High order correlation model for associative memory. In Aip conference proceedings (Vol. 151, pp. 86–99), 1986.